\documentclass[conference,a4paper]{IEEEtran}
\IEEEoverridecommandlockouts

\usepackage{graphics} 
\usepackage{epsfig} 
\usepackage{mathptmx} 
\usepackage{times} 
\usepackage{amsmath} 
\usepackage{amssymb}  
\title{\LARGE \bf
Dynamic Routing for Traffic Flow through Multi-agent Systems
}

\begin{document}

\author{
\IEEEauthorblockN{Jizhe Zhou}
\IEEEauthorblockA{\textit{Department of Computer and Information} \\
\textit{Science, University of Macao}\\
Macao, China \\
yb87409@um.edu.mo}
\and
\IEEEauthorblockN{Qiwei Chen}
\IEEEauthorblockA{\textit{Wujing Technology Co. Ltd.} \\
{Zhuhai ,China}}
\and
\IEEEauthorblockN{Qin Li}
\IEEEauthorblockA{\textit{East Chine Normal University} \\
Shanghai, China}
}

\maketitle
\thispagestyle{empty}
\pagestyle{empty}


\begin{abstract}
Routing strategies for traffics and vehicles have been historically studied. However, in the absence of considering drivers' preferences, current route planning algorithms are developed under ideal situations where all drivers are expected to behave rationally and properly. Especially, for jumbled urban road networks, drivers' actual routing strategies deteriorated to a series of empirical and selfish decisions that result in congestion. Self-evidently, if minimum mobility can be kept, traffic congestion is avoidable by traffic load dispersing. In this paper, we establish a novel dynamic routing method catering drivers' preferences and retaining maximum traffic mobility simultaneously through multi-agent systems (MAS). Modeling human-drivers' behavior through agents' dynamics, MAS can analyze the global behavior of the entire traffic flow. Therefore, regarding agents as particles in smoothed particles hydrodynamics (SPH), we can enforce the traffic flow to behave like a real flow.  Thereby, with the characteristic of distributing itself uniformly in road networks, our dynamic routing method realizes traffic load balancing without violating the individual time-saving motivation. Moreover, as a discrete control mechanism, our method is robust to chaos meaning driver's disobedience can be tolerated. As controlled by SPH based density, the only intelligent transportation system (ITS) we require is the location-based service (LBS). A mathematical proof is accomplished to scrutinize the stability of the proposed control law. Also, multiple testing cases are built to verify the effectiveness of the proposed dynamic routing algorithm.

\end{abstract}

\section{INTRODUCTION}
\par 
\begin{figure*}[tp]
      \centering
      \includegraphics[scale=0.5]{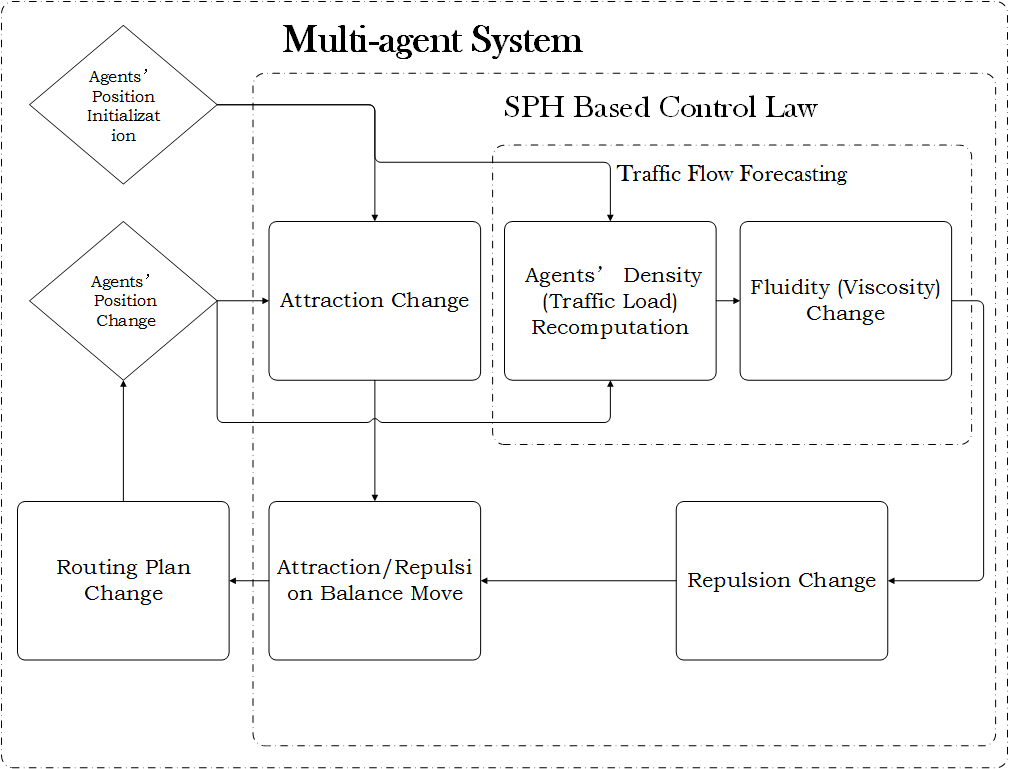}
      \caption{Conceptual working diagram of our MAS based, SPH styled dynamic routing method}
      \label{figurelabel}
   \end{figure*}
\par 

\par 
Traffic routing is a well-established research and optimization problem in traffic management \cite{c21}. Most research has been done for static problems, that is, settings where the problem structure does not change. In static problems the routing decision boils down to finding the shortest path between the origin and the destination point. Once a solution has been found for all routes, the optimal ones can be used whenever needed. These algorithms typically are based on shortest path algorithms, like the well-known $A^*$ algorithm.

\par 
The situation becomes more complex if we regard dynamic problems. In a dynamic problem, the problem structure changes while solving the problem. For routing decisions this implies that it is not the traveling distance that has to be optimized but the traveling time. Of course, a simple approach is to assume a fixed average speed that can be used for every road and to use this in the calculation of the weights of the road network. It has turned out that this simple approach often can help, but this also turned out not to be sufficient for roads whose load varies with time. Heavy traffic is typical for urban areas nowadays. However, with another observation in these areas, there often exist a number of alternative routes, as well. In this article, we investigate how to handle dynamic routing problems in urban areas, with high traffic and a number of alternative routes.

\par 
Applying current trends and technologies like car-to-car communication\cite{22} and autonomic road transportation support systems \cite{23}, cars can be enabled to communicate with each other, and also with their environment. Therefore, each car can be seen as an autonomous entity, which has computing abilities and is able to send and receive information from its current environment and to use this information. One class of routing algorithms that can take advantage of these latest developed technologies to provide routing in dynamic environments is the multi-agent system (MAS)based optimization approaches. To satisfy drivers' motivations as well as maximize the benefit of the entire traffic flow, the multi-agent system (MAS) is introduced as the coarse-to-fine framework. MAS is a computerized system composed of multiple interacting intelligent agents and used to model a complex system by decomposing it in small entities (agents) and by designing actions on individual level \cite{c1}. Our MAS framework is employed to track travelers’ behaviors on the road and develop the basic traffic flow model through a decomposed structure. Therefore, MAS helps us migrate the original traffic routing problem to a MAS control problem.

Despite the use of multi-agent systems, as heavy traffic is typical for urban areas nowadays, building a proper control algorithm to achieve the optimization of routes for all travelers with the same destination is extremely computationally complex, taking into account the possibility of all routes. Nevertheless, it is difficult to solve analytical systems with physics tools, the similarities between fluids and routing under the above cases are easy to be understood. Thereby we investigate and establish our dynamics according to the famous smoothed particles hydrodynamics (SPH) algorithm among agents to minimize congestion. 
\par
SPH is a computational method in fluid mechanics used for simulating the dynamics of incompressible continuum media that varies from solid to fluid or even gas \cite{c7}. Mainly, we add gravity like attraction and fluid pressure based repulsion among agents to minimize congestion. The attraction encourages travelers to choose the fastest route. Repulsion penalizes their choices of the same route under heavy traffic load. Besides, a built-in traffic flow forecasting (TFF) strategy is constructed by utilizing the viscosity term in the repulsion. Fluids tend to distribute themselves uniformly within a space. In other words, as the \textbf{Pascal's principle} stated, forces can be conducted within liquids. Therefore, dense regions with a high dissipative viscosity force could conduct this part of repulsion to the rest part of the fluid. With distance weighted SPH kernel function, viscosity will significantly affect the routing decisions of upcoming agents and slightly affect the agents at a distance.  

In such a  way, we transformed the dynamic routing problem to a MAS control problem which the MAS behaves like a rare fluid with a varies viscosity. Moreover, we finished a stability analysis for our MAS based dynamic routing system through constructing a proper Lyapunov function.
As stated above, the major contributions of our work are:
\begin{itemize}
\item We build a de-centralized, multi-agent based dynamic routing strategy that has much higher fault-tolerance, adaptive, and computation efficiency without the requirement of global information while comparing to currently applied centralized routing strategies.
\item We transform the routing problem into the fluidity control problem in MAS that can satisfy the drivers' preferences from individual perspective as well as optimize the global traffic load balance on alternative routes.
\item We proved the convergence of our system by finding a Lyapunov stability, and this is extremely meaningful for adapting our system to real applications.    
\end{itemize}

\par
The working pipeline of the proposed dynamic routing method is shown in Figure 1.

\section{RELATED WORK}
\par
Path planning is not a new topic under the traffic flow control field. Earlier than 1990s, researchers have already built a complete series of theory for this topic\cite{c2,c3}. However, these early proposed frameworks are mostly centralized which demands a central computer to realize the global information summation and overall vehicles dispatching.

Generally, the simulations of traffic flows could be set into micro, medium and macro scope\cite{c17}. Each of them are standing at different level of scope in simulation\cite{c4}. Naturally, the macroscopic traffic flow model arises just within a few decades is a very clear result for decentralized control\cite{c18}. Instead of focusing on individual vehicles, the macroscopic model considers the average behavior of a entire flow and generates both time and space continuous mathematical expression\cite{c4}.

The macroscopic model treats traffic flow as some kind of incompressible fluid flow. And the initially applied fluid mechanics in simulating traffic model is in the Euler perspective. Euler perspective is mainly allocated within the cell transmission model (CTM) model \cite{c5}. However, to deduce a continuous function of flow, the Lagrangian perspective is a must. Therefore, SPH, the representative of Lagrangian perspective is selected. 
\par 
Latest works focused on obstacle avoiding in robotics control area have already noticed such two algorithms \cite{c6,c7,c13}. The state-of-the-art work\cite{c7} designed a feedback control law for swarms of robots based on SPH algorithms. This decentralized feedback controller is built under the auxiliary of SPH mechanism. Agents are forced to act similarly as particles in a fluid from the global view. Parameters such as pressure and external forces are carefully revised in order to coordinate the agents properly. The external forces are omitted and superseded by a harmonic global potential function under most cases. Some similar works on robotics control are\cite{c8,c19}. To our best knowledge, we are the first to conduct research on traffic flow control according to MAS and SPH.
\section{Basic Smoothed Particle Hydrodynamics}
SPH is an interpolating algorithm that each particle carries its own attributes and interacts with its neighbors. 
A detailed description for computer simulation algorithm of SPH is introduced by Muller \cite{c9}. Although different SPH versions existed, they are all considered equivalent according to \cite{c10}. 
\par
Standing at the Lagrange perspective in fluid mechanics, SPH realized a mesh-free way to simulate continuous fluid's behavior. The continuous fluid is decomposed into particles, and these particles interact with each other to form the complex fluid properties ultimately. Moreover, for every single particle, it obeys Newton's second law and SPH is built on the integral function:
\begin{equation}  
f(q)= \int_{\lambda} f(q') \delta (q-q')dq' 
\end{equation}  
$\lambda$ is the volume that contains a particle located at position $q$. $\delta(q-q')$ could be regarded as the Dirac function which decides the communication range $D$ for each agent/vehicle. $f(q)$ could be any attribute we want to compute at location $q$.
\begin{equation}  
\delta(q)= \left\{
\begin{array}{rcl}
q \leq D, 1 \\
q > D, 0
\end{array}
\right.
\end{equation}  
\begin{figure}[bhtp]
      \centering
     
      \includegraphics[scale=0.5]{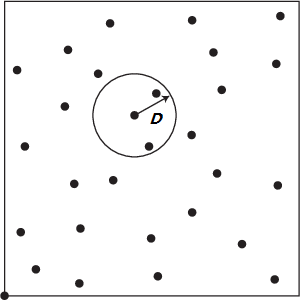}
      \caption{Communication range of agents}
      \label{figurelabel}
   \end{figure}
Substituting the Dirac function $\delta$ by a smoothing kernel function $W$ to display the different effects of neighbor particles based on different distance and $W$ satisfies the conditions for stabilized SPH. This kernel function sticks the communication range in multi-agent systems and the smoothing kernel of SPH together. We can get:
\begin{equation}  
f(q)= \int_{\lambda} f(q') W(q-q',h)dq' 
\end{equation}
Here, the $W$ function is smoothing kernel function and $h$ is the smoothing length relatively.

\par
Considering a particle $j$ has a finite volume $\Delta V_{j}$ in (3), this integration becomes a 
summation function that counts total $N$ neighbor particles' effect of particle $i$ that located at position $q$.
\begin{equation}  
\Delta V_{j}=\frac{m_{j}}{\rho_{j}}
\end{equation}
Thus, the integration could be rewrite into summation format by replacing $\Delta V_{j}$ ($\Delta V_{j}$ may not equal to $\lambda$):
\begin{equation}  
f(q)\approx \sum_{j=1}^{N} \frac{m_{j}}{\rho_{j}} f(q_{j}) W(q-q_{j},h)
\end{equation}
Additionally:
\begin{equation}  
\left\{  
             \begin{array}{lr}  
             \frac{D\rho}{Dt} = - \rho \nabla \cdot v \\
             \\
             \frac{Dv}{Dt} = - \frac{\nabla P}{\rho} \\
             \\
             \frac{De}{Dt} = - (\frac{P}{\rho}) \nabla \cdot v\\

             \end{array}  
\right.  
\end{equation}  
For general SPH based robotics control strategies, the “three continuum governing equations of fluid dynamics" are applied and deduced from above summation equation.

\section{Dynamic Routing Method}
Equation set (6) is the fundamental of many related works. In \cite{c6}, their entire SPH based control law is derived from (6). However, since the SPH algorithm serves the multi-agent systems as its control law, the pure SPH model cannot meet our requirements completely.Moreover, as the SPH algorithm must comply with Newton's Second Law, we revised a novel sophisticated SPH model for our control problem. 
Based on the premise that every agent/particle in our system shares an exactly same unit volume $\Delta V$, the following is archived by utilizing the Newton's Second Law:
\begin{equation}
m \vec a=\vec f  \Rightarrow    \rho \vec a =\vec f
\end{equation}
$\Delta V$ is omitted as a unit and $\vec f$ actually means the fundamental unit of a force. The  total force for a single agent is made up of 3 terms according to SPH.
\begin{equation}
\vec f^{total}=\vec f^{external}+\vec f^{pressure}+\vec f^{viscosity}
\end{equation}
$\vec f^{external}$ represents the external forces on an agent, commonly the gravity.
\begin{equation}
\vec f^{external}=\rho \vec g
\end{equation}
$\vec f^{pressure}$ is brought in by pressure differences inside the fluid. Numerically, it equals to the gradient of the pressure field and points from high pressure region to the lower one.
\begin{equation}
\vec f^{pressure}=- \nabla P
\end{equation}
$\vec f^{viscosity}$ is generated by the velocity differences among particles. The faster particles will apply a shear force to the slower ones. Under some certain circumstance, the bulk force also needs to be considered. Thus, the viscosity force always acts the way that pushes the slower particles to move faster or prevent the faster particles from segregation. In early works of multi-agent systems or swarm robotics, this $\vec f^{viscosity}$ term is purely treated as an artificial viscosity term that could be ignored with no harm. Some very latest research has revealed the usage of viscosity force in adjusting the global behavior of SPH based multi-agent systems \cite{c11}. In order to merely instruct the basic control framework, we follow the concrete definition of viscosity and write it as a coefficient related to the velocity difference here:
\begin{equation}
\vec f^{viscosity} = \mu {\nabla}^2 v
\end{equation}
Above $\mu$ is the so called artificial viscosity coefficient commonly set between $(0,1)$. Replace forces in (8) with densities in (9) (10) (11):
\begin{equation}
\rho \vec \alpha = \rho g + (-\nabla P) + \mu {\nabla}^2 v
\end{equation}
\par 
The above equation (12) is a simplified version of the famous Navier-Stokes equation\cite{c10}. (12) is the basic framework to design the control law of our traffic flows since a traffic flow is considered as a kind of real physical flow. In the following chapters, we introduce series of adaptations based on (12) to model and navigate the traffic flow as required.

\section{Adopted Smoothed Particle Hydrodynamics}
As we stated above, the generalized control law should be:
\begin{equation}
\ddot{q_{i}}= u_{i}
\end{equation}
Where $\ddot{q_{i}}$ is the acceleration for agents, and $u_{i}$ is the SPH formatted control input.
Consciously knowing the shape of dynamics in a multi-agent system, we choose to use the much closer total forces representation of the SPH algorithm in (12) as the basic control law.
\par
Let $\vec a$ in (12) matches $\ddot{q_{i}}$ in (13), we can get (14) as:
\begin{equation}
\ddot{q_{i}} \equiv \vec a_{i}= g + \frac{(-\nabla P_{i})}{\rho_{i}} + \frac{\mu {\nabla}^2 v_{i}}{\rho_{i}}
\end{equation}
According to Newton's Second law, in (14), $\vec a_{i} = (\rho_{i}^{-1} \cdot \vec f^{total}_{i})$.  To make up for the gap, we fusion the $\rho^{-1}_{i}$ term into $\vec f^{total}_{i}$ during computation. But mathematically, the Newton's Second Law still reserves. 
Thus the control law becomes:
\begin{equation}
\ddot{q_{i}} \equiv \vec a_{i}\equiv \vec f^{total}_{i}\cdot \rho^{-1}
\end{equation}

\subsection{Global Potential Functions}

\par
To model and navigate the traffic flow on the road, we should know the road first. Overall, roads in a multi-agent navigation problem are considered as a 2D model with length and width. Following the idea of regarding traffic flow like a real physical flow and inspired by the widely applied artificial potential fields methods\cite{c20,c21}, we revise the roads as a 3D model with extra height dimension. The destination acts as the gravity offering large attraction power to travelers, and each traveler will try driving at their best speed to reach the destination without interference. As far as we know, the roads never gain altitude attribute in any former works.
\par 
The existing of detours leads to various choices of routes, referring to potential field method, as shown in Fig. 4., we segment the road when each branch (junction )starts


   \begin{figure}[bhtp]
      \centering
      \includegraphics[scale=0.35]{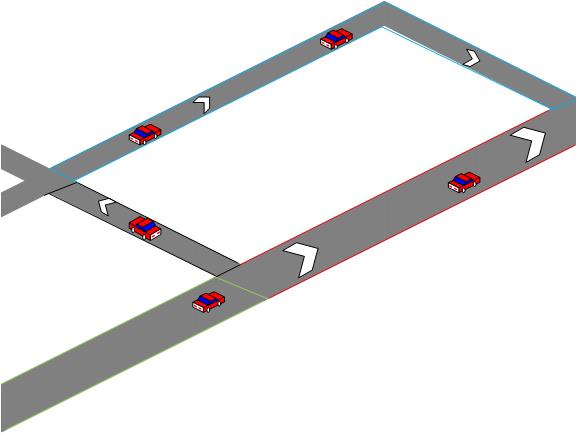}
      \caption{Segmentation of roads}
      \label{figurelabel}
   \end{figure}
\par 

\par
After segmentation, we can view the topology of the total routes as road-map. Maintaining the longitude of each segmentation, 2D road segmentation arose to slopes as follows. 
\begin{itemize}
\item The starting location $q_{start}$ is a constant for travelers stem from same locations and the final destination $q_{end}$ is a constant for all travelers. $Dis=\Vert(q_{end}-q_{start})\Vert$ decides the straight line distance between two locations. Therefore, $Dis$ is also a constant to travelers start at same locations in our model.
\item Once an agent picks to travel along a certain segment of a road $R_{n}$ with length $L_{(R_{n})}$, it must bring the agent some distance $Dis_{(R_{n})}$ straight forward or close to the $end$ location, i.e. $Dis=\sum_{n} Dis_{(R_{n})}$. Noticing that commonly length $L_{(R_{n})} \gg Dis_{(R_{n})}$.
\item Therefore, we can raise the $R_{n}$ road segmentation to a slope with an angle $\theta=\arcsin({Dis_{(R_{n})}}/{L_{(R_{n})}})$
\end{itemize}
\par
Manifestly, the $Dis_{(R_{n})}$ term is much smaller than $L_{(R_{n})}$ in numeric value since it stands for straight line distance. Not explicit state in above, we always segment the roads as less as possible to avoid meaningless computation. In other words, showing in Fig.3., the segment operation only happens at intersections of roads.
\par
Thus, the gravity term in our SPH control algorithm will be adopted to:
\begin{equation}
\vec f^{external}= \rho g \sin\theta
\end{equation}

 \begin{figure}[bpht]
      \centering
     
      \includegraphics[scale=0.5]{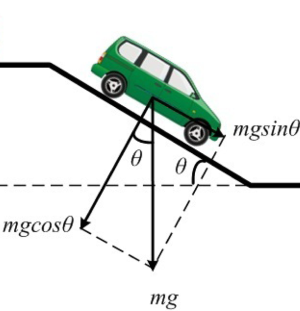}
      \caption{Simulation pipeline}
      \label{figurelabel}
   \end{figure}
\par

As we specified at the beginning of this subsection, the global potential function is introduced in path planning to solve the convergence problem in many previous works. Obeying such settings, we also defined a global potential function $\Phi(q_{i})$. Therefore the negative gradient of this function should equal to (17):
\begin{equation}
\nabla\Phi(q_{i})=
\left\{
\begin{array}{lr}

- g \sin{\theta} \quad if \quad q_(i) \neq q_{end} \\
\\
0  \quad if \quad q_(i) = q_{end}

\end{array}
\right.
\end{equation}
\par
Notice that $\vec a^{external}_{i} =\vec f^{external}_{i} \cdot \rho^{-1}$, and the global potential function computes the acceleration of a vehicle directly through the negative gradient.

\subsection{Abridged Pressure Controller}
\par
For the $\vec f^{pressure}$ force, we compute the standard SPH algorithm to derive its density equation.
\begin{equation}
\vec f_{i}^{pressure}=- \nabla P_{i}   = -\sum_{j}m_{j}\frac{P_{j}}{\rho_{i}\rho_{j}}\nabla_{i}W_{ij}
\end{equation}
To shorten our equations, $W_{ij}=W(q_{i}-q_{j})$, with kernel function $W$. The above pressure equation is mathematically right but not physical meaningful since it is not symmetric and will cause unequal reaction. This will violate Newton's Third Law.

The popular way for symmetrization is:
\begin{equation}
\vec f_{i}^{pressure}= -\sum_{j} m_{j} (\frac{P_{i}}{\rho _{i}}+\frac{P_{j}}{\rho _{j}}) \nabla_{j}W_{ij}
\end{equation}
$P_{i}/\rho_{i}$ computes the current time $t$'s status within the communication range $D$ of agent $i$. 
$P_{j}/\rho_{j}$ predicts the movement in next control cycle $(t+\Delta t)$ within the $2D$ range of agent $i$. Equation (19) is the applied pressure term for SPH controller in many related multi-agent systems\cite{c11,c12}. 

Critically, in our model, the agents are trying to move through a high-density background and distributed uniformly. Considering the symmetric kernel function $W$, the integral of its first derivative is zero. Thus, when a large number of vehicles are driving on the road, the following could be attempted:
\begin{equation}
\sum_{j} \frac{P_{j}}{\rho _{j}} \nabla_{j}W_{ij} \approx 0
\end{equation}
Then, the abridged $\vec f_{i}^{pressure}$ is:
\begin{equation}
\begin{array}{lr}
\vec f_{i}^{pressure}= -(\frac{P_{i}}{\rho _{i}}) \sum_{j} m_{j}\nabla_{j}W_{ij}\\
\\
\vec a_{i}^{pressure}= -(\frac{P_{i}}{\rho _{i}^2}) \sum_{j} m_{j}\nabla_{j}W_{ij}
\end{array}
\end{equation}

However, (21) is not enclosed since agents only carry three initial quantities: mass, position, and velocity. The pressure is not a characteristic of agents. It changes along the agents' location. Thus, the pressure needs to be evaluated firstly at every time step. Well aware that our traffic flow is physically incompressible, the pressure can be derived from density through the ideal gas state equation. Although named as gas, the ideal gas state equation always functions on fluid mechanics field.
\begin{equation}
P_{i} = k \rho_{i} \quad k \in (0,1)
\end{equation}
In this paper, we used this modified version of (22) due to traffic flows are highly deformable.
\begin{equation}
P_{i} = k (\rho_{i}-\rho_{rest})
\end{equation}
$\rho_{rest}$ is the rest density. Since pressure forces depend on the gradient of the pressure field, the offset does not affect pressure forces. However, the offset does influence the gradient of a field smoothed by SPH and makes the simulation numerically more stable\cite{c10}.

\subsection{Density Controlled Kinematic Viscosity}
\par
Similar to widely introduced 2-layer control law in multi-agent/swarm robotics system, the projected gravity $\vec f^{external}$ offers the global potential field styled attraction to drive the vehicles directly towards the destination. The second $\vec f^{pressure}$ term provides a relative repel force to each other. The pressure is conductive in the fluid, and press the particles from the high-density area to the low-density area. The remaining viscosity force which is neglected in earlier research on this field acts as the navigation strategy for each vehicle in our model. Imitating the pressure part which is considered to be the main contribution of SPH algorithm, our viscosity force should be conductive, sensitive to density and significant in changing agents behavior. The recent studies on SPH based multi-agent systems already valued the importance of viscosity. The commonly employed variation is given by \cite{c13}.
\par
In state-of-the-art work \cite{c7}, the dynamic viscosity $\mu$ of a fluid expresses its resistance to shearing flows, where adjacent layers move parallel to each other with different speeds. $\mu$ cooperate with average densities, viscosity constant and the speed of sound to damping the agents from penetration. These works still follow the basic artificial viscosity assumption of SPH where viscosity coefficients is a constant used to stabilize the whole system without making salient behavior change to particles. 
\par
Regarding viscosity term is the most critical tuning item for navigation, kinematic viscosity is introduced. Dividing the dynamic viscosity by density, we get the kinematic viscosity $\nu$ which is strongly related to the Reynolds number $Re$.
\begin{equation}
\nu_{i} = \mu_{i} / \rho_{i} \\
\end{equation}
\par 
Reynolds number describes the feature the flow's movements. Let $\nu$, $\rho$ , $v$ stands for the stacked average kinematic viscosity, density and velocity on a $L$ long road for all agents:
\begin{equation}
Re= \rho v L/ \nu
\end{equation}
\par 
In standard SPH algorithm for fluids, especially water, the entire flow is designed as an incompressible, zero viscous, idea flow since its density and viscosity are not allowed to change with time. In our model, applying the kinematic viscosity that changing with time is the core contribution of our feedback control law to navigate the agents until balancing. As the scenario proposed at the beginning of this work, travelers are seeking the best travel time under heavy traffic load on the road. Making the drivers willing to abandon partial vested interests and detour rationally to shorten travel time is the task of our dynamic routing algorithm. Thus, we use the kinematic viscosity based routing strategy to adjust the flexibility of particles. But, a varying $\nu$ in the (25) makes $Re'$ hard to describe according to (24).
\par 
The high density must lead to viscous appearance and slow speed of flow. Moreover, working as a navigation system for dynamic path planning, a smaller Reynolds number which refers the viscosity is significant enough in changing the fluidity of the flow is always preferred in our system. $L$ and $\rho$ are constant at the global view, the small enough $Re$ shows viscosity generates a significant resistance force in contrast with the gravity/attraction generated velocity or known as the \textbf{Laminar Flow}. Avoiding introducing additional new variables into the system, we reform $\mu$ by imitating the temperature effect in viscoelastic fluids:
\begin{equation}
\log (\log (\mu_{i}+\gamma))=a-b\log(\rho_{i}^{-1})
\end{equation}
\par
$\mu$ is the desired viscosity, $\gamma \in [0.6,0.9]$, and $a$,$b$ are positive constants related to the agents' materials. Therefore, according to (11) the $\vec f^{viscosity}$ could be formed as \cite{c29} proposed. But the most commonly used artificial viscosity is deduced from the momentum conservation equation in (6):
\begin{equation}
\begin{array}{lr}
\vec f^{viscosity}_{i}= \frac{\rho_{i}}{m_{i}}\sum_{j}\lbrack ({m_{j}}) (\mu_{i}+\mu_{j}) (\frac{v_{ij}q_{ij}}{\Vert{q_{ij}}\Vert ^ {2}+\eta ^2} \nabla_{j} W_{ij})   \rbrack \\
\\
\vec a^{viscosity}_{i}= \frac{1}{m_{i}}\sum_{j}\lbrack ({m_{j}}) (\mu_{i}+\mu_{j}) (\frac{v_{ij}q_{ij}}{\Vert{q_{ij}}\Vert ^ {2}+\eta ^2} \nabla_{j} W_{ij})   \rbrack
\end{array}
\end{equation}
$q_{ij}$ is the abbreviation for $(q_{i}-q_{j})$ to reduce the length of our equation. Similar to pressure force, the viscosity generated acceleration in (27) is not symmetrical. A popular viscosity form is given in \cite{c10}.
\begin{equation}
\vec a^{viscosity}_{i}= 
\left\lbrace
\begin{array}{lr}
\sum_{j}m_{j}\{-[(\mu_{i}+\mu_{j}) \frac{v_{ij}q_{ij}}{\Vert{q_{ij}}\Vert ^ {2}+\eta ^2}]\\
+[(\mu_{i}+\mu_{j})^2 \frac{v_{ij}q_{ij}}{\Vert{q_{ij}}\Vert ^ {2}+\eta ^2}]^2\}, \\
\qquad if \quad v_{ij}\cdot q_{ij} < 0\\
\\
0, \qquad if \quad v_{ij}\cdot q_{ij} > 0
\end{array}
\right.
\end{equation} 
$v_{ij}$ is also the abbreviation for $(v_{i}-v_{j})$ and $\eta ^2$ is the Laplace smoothing term introduced to avoid singularities. The viscosity in (28) vanishes when $v_{ij}\cdot q_{ij} \ge 0$, which is the SPH equivalent of the condition $\nabla \cdot v \ge 0$. Only when the stacked velocity decreases or the later agents tend to overtake the place of former ones', the viscosity functions. A lot more properties about (28) could be further found in \cite{c10}.
\par
In (24), $\nu$ is associated with $\rho$, we can derive:
\begin{equation}
\nu_{i} \propto e^{\rho_{i} ^b} \quad (b \in R^{+})
\end{equation}
Since $\rho_{i}$ is the number of agents circling agent $i$, $\rho_{i}$ belongs to $N^+$. Thus, $\nu_{i}$ increases exponentially as $\rho_{i}$ grows. Placing (28) into (14), $Re$ decreases very fast when $\rho$ increasing. A relatively low Reynolds number will be generated on the high-density fields of a flow. Alternatively, the low-density regions influenced more by attraction/gravity that accelerates the agents to the destination.

\par
Reexamining our model under total force format:
\begin{equation}
\begin{array}{lr}
\vec f^{total}_{i}=\vec f^{external}_{i}+\vec f^{pressure}_{i}+\vec f^{viscosity}_{i}\\
\\
\vec a^{total}_{i}=\rho^{-1} \vec f^{total}_{i}=\vec a^{external}_{i}+\vec a^{pressure}_{i}+\vec a^{viscosity}_{i}
\end{array}
\end{equation}
Substituting the right side forces in (8) with (16),\quad(21),\quad(28), our numerical solver of the Navier-Stokes equation is derived. 
\par 
Our controller is derived by considering a vehicle as an SPH particle at location $q_{i}=[x_{i},y_{i}]^{T}$ subjected to the total force $\vec f^{total}_{i}$. In this model, the second-order dynamics of vehicles is applied.$-\xi v_{i}$ is the dissipative term with a positive coefficient of $\xi$ set to help stabilize the whole system. We will prove the damping functionality of $-\xi v_{i}$ later. $\xi= 0.3$ is applied in this work. Based on former assumptions, each vehicle's acceleration is given by:
\begin{equation}
\ddot{q_{i}}= u_{i}=\vec a^{total}_{i}- \xi v_{i}
\end{equation}
\par
According to (30) and (31), the drivers' preferences are always satisfied. The current routing strategy always points to the lowest density area that offers drivers the highest travel speed. 
\par

\section{Stability Analysis}
\subsection{Stability and Convergence Analysis}
In \cite{c14, c15}, they have proposed some significant effective ways in analyzing the stability of SPH based multi-agent systems with linear viscosity force. Inspired by them and combine some outstanding skills presented in \cite{c16}, we offered an analysis of stability on our model with non-linear navigational feedback control law.
\par
We define a system level function $\Phi$ to measure the overall performances of our control law:
\begin{equation}
\Phi_{S}(\bar{q})= \sum_{i} \Phi (q_{i})
\end{equation}
Where $\Phi (q_{i})$ is the global potential function we defined in (17). Based on $\vec f^{external}$ that acts as a global potential function, $\Phi_{S}$ provides the measurements of how close the agent is to the destination $q_{end}$. Meaning greater $\Phi_{S}$ leads to a greater distance for all agents to their common destination. Therefore, $\Phi_{S}$ has a unique minimal point $\Phi_{S}=0$ at $q_{end}$. 
\par
The minimum value of $\Phi_{S}=0$ is archived when all agents reached $q_{end}$.
Therefore the primary goal of the control law is to minimize $\Phi_{S}$. Consider $\vec f^{external}$, the Hessian of $\Phi_{S}, H_{\Phi_{S}}$ should equal to $0$ when agents already arrived the $q_{end}$. $\vec f^{external}=0$ leads to $\Phi_{S}=0$ since $0$ distance is left to destination. And for our designed $\theta \in [0,\frac{\pi}{2}]$, $\int (-\sin\theta) =\cos\theta \ge 0$. Reasonably, $\Phi_{S} \ge 0$. In all, we can conclude that $\Phi_{S}$ is positive semidefinite. 
\par
Consider a Lyapunov function as below:
\begin{equation}
V(q)= \Phi_{S}(\bar q)+\sum_{i} \dot{e_{i}} +\sum_{i}\frac{1}{2}v^{T}_{i}v_{i} 
\end{equation}
$\dot{e_{i}}$ is the energy partial to time which is given in (6) as the energy conservation equation. Obviously, $V(q)$ is positive semidefinite since $V(q)=0$ when all $v_{i}=0$ meaning $\Phi_{S}=0$. And $V(q) \ge 0 \quad \forall q$. Differentiate (33) respect to time $t$ and $\Phi(q_{i})=\Phi_{i}$ for simplification, we can get:
\begin{equation}
\begin{aligned}
&\dot{V}(q)=\sum_{i}(\nabla \Phi^{T}_{i} (\dot{q_{i}})+v^{T}_{i}\dot{v_{i}}+  \frac{\partial \dot{e_{i}}}{\partial t} )\\
\Rightarrow &\dot{V}(q)=\sum_{i}(\nabla \Phi^{T}_{i}v_{i}+v^{T}_{i}u_{i}+ \frac{\partial \dot{e{i}}}{\partial t}   )\\
\Rightarrow &\dot{V}(q)=\sum_{i}(\nabla \Phi^{T}_{i}v_{i}+v^{T}_{i}(\vec a^{total}-\xi v_{i})+ \frac{\partial \dot{e{i}}}{\partial t}   )\\
\Rightarrow&\dot{V}(q)=\sum_{i}( \nabla \Phi^{T}_{i}v_{i})+\sum_{i} v^{T}_{i}\{-\sum_{j} m_{j} (\frac{P_{i}}{\rho ^{2}_{i}}) \nabla_{j}W_{ij}\\
&+\vec a^{viscosity}_{i}-\xi v_{i} +g \sin \theta)\}+\sum_{i} \frac{\partial \dot{e_{i}}}{\partial t}
\end{aligned}
\end{equation}
Replace (17) in (34) and let $v_{ij}=v_{i}-v_{j}$ to denote the velocity difference matrix as gradient of $v$ in (6). We got $\frac{\partial\dot{e_{i}}}{\partial t}= \frac{1}{2}\sum_{j}\frac{P_{i}}{\rho_{i}^2}v_{ij}\nabla_{i}W_{ij}  $ and
\begin{equation}
\begin{array}{lr}
\dot{V}(q)=\sum_{i}( \nabla \Phi^{T}_{i}v_{i})+\sum_{i}v^{T}_{i}(-\nabla\Phi_{i})+\\
\\
\sum_{i} v^{T}_{i}\{-\sum_{j} m_{j} (\frac{P_{i}}{\rho ^{2}_{i}}) \nabla_{j}W_{ij}+\vec a^{viscosity}_{i}

-\xi v_{i}\} + \\
\\
\sum_{i} \frac{1}{2} \sum_{j} m_{j}(\frac{P_{i}}{\rho^{2}_{i}})v^{T}_{ij}\nabla_{i}W_{ij}
\end{array}
\end{equation}
\par
We can further simplify that $\nabla_{i}W_{ij}=-\nabla_{j}W_{ji}$, $v_{ij}=v_{i}-v_{j}$ and $m_{i}=m_{j}$, where $m_{i}\neq m_{j}$ for different vehicles in real simulation. Set $\sum_{j}m_{j}(\prod_{ij})\nabla_{i}W_{ij}=\vec a_{i}^{viscosity}$ in (34) to shorten our computation. Since viscosity is also symmetrized $\prod_{ij}=\prod_{ji}$. (35) could be further deducted as:
\begin{equation}
\begin{aligned}
&\dot{V}(q)=\sum_{i} v^{T}_{i}\{-\sum_{j} m_{j} (\frac{P_{i}}{\rho ^{2}_{i}} +\prod_{ij})\nabla_{i}W_{ij}-\xi v_{i}\}\\
&+\sum_{i}\frac{1}{2}v^{T}_{i}\sum_{j}m_{j}(\frac{P_{i}}{\rho^{2}_{i}})\nabla _{i}W_{ij}
+\sum_{j}\frac{1}{2}v^{T}_{j}\sum_{j}m_{i}(\frac{P_{i}}{\rho^{2}_{i}})\nabla _{j}W_{ji}\\
\Rightarrow &\dot{V}(q)=-\sum_{i} \xi v^{T}_{i}v_{i}
-\sum_{i}\frac{1}{2}v^{T}_{i}\sum_{j}m_{j}\prod_{ij}\nabla _{i}W_{ij}\\
&-\sum_{j}\frac{1}
{2}v^{T}_{j}\sum_{j}m_{i}\prod_{ji}\nabla _{j}W_{ji}\\
\Rightarrow &\dot{V}(q)=-\sum_{i} \xi v^{T}_{i}v_{i}-
-\sum_{i}\frac{1}{2}\sum_{j}m_{j}\prod_{ij}v_{ij}^T\nabla_{i}W_{ij}
\end{aligned}
\end{equation}
\par

(36) seems to be a mess. However, if we take a closer look at the viscosity generated acceleration, $\vec a^{viscosity}_{i}$ only exists within (36). The SPH smoothing kernel has the characteristic\cite{10}:
\begin{equation}
\nabla_{i}W_{ij}=-\Vert \nabla_{i}W_{ij} \Vert \frac{q_{ij}}{\Vert q_{ij} \vert}
\end{equation}
And from (28) we can know the fact that $\vec a^{viscosity}_{i} > 0$ when $v^{T}_{ij}q_{ij} < 0$. Meaning $\prod_{ij}>0$ when $v^{T}_{ij}q_{ij} < 0$. Thus place (37) into (36):
\begin{equation}
\begin{array}{lr}
\dot{V}(q)=-\sum_{i} \xi \Vert v_{i} \Vert ^2-\\
\\
\sum_{i}\frac{1}{2}\sum_{j}m_{j} \prod_{ij} \Vert \nabla_{i}W_{ij} \Vert \frac{-v^{T}_{ij}q_{ij}}{\Vert q_{ij} \vert} \leq 0
\end{array}
\end{equation}
Therefore, from the \textbf{LaShelle's Invariance Principle}, we conclude that the solutions of the entire system starting form $\Omega_{C}$ will converge to the largest invariant set $\Omega_{I}$ that  $\Omega_{I} \in \{x \in X| \dot{V{q}}=0       \} $ and $v_{i}=0, \forall i$. Thus our system will get a equilibrium state when all agents remains statical. Meaning either all agents reached their destination or all possible routes are overloaded and unavoidable congestion happens.

\section{Simulation and Discussion}

To illustrate the performance of our model under such cases, we use a cloverleaf interchange scenario. This scenario is diverse enough to reveal all the causes for congestions. Under the help of $\boldmath Anylogic$, a cloverleaf is set up.

\begin{figure}[bpth]
      \centering
     
      \includegraphics[scale=0.25]{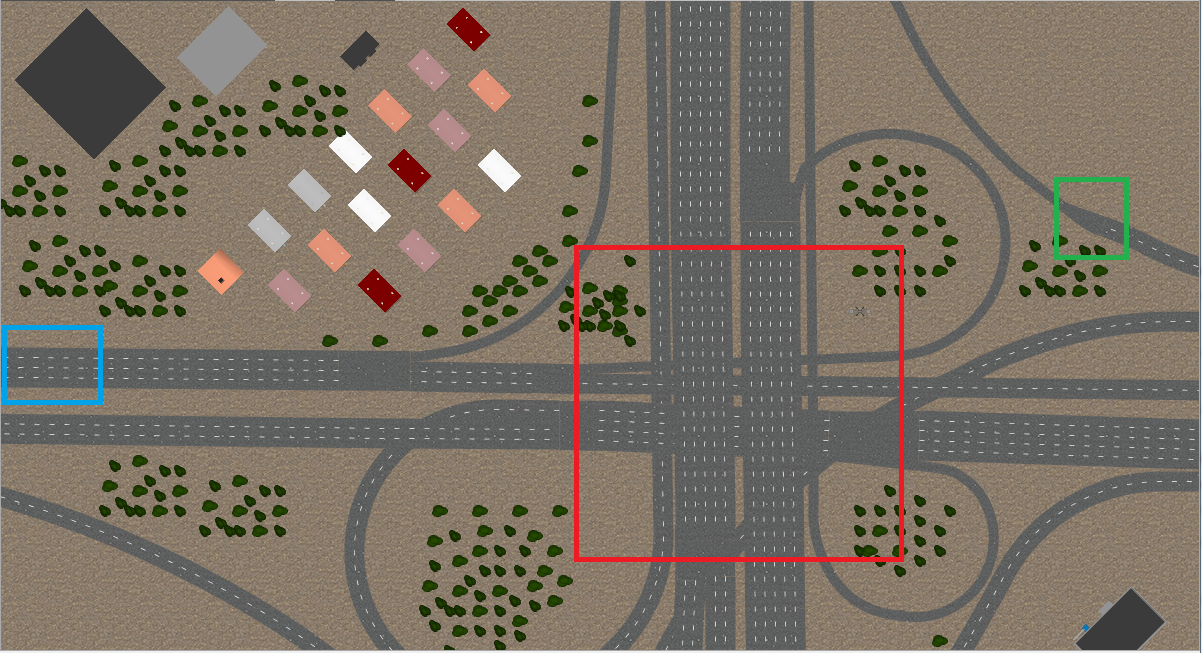}
      \caption{The cloverleaf interchange system}
      \label{figurelabel}
   \end{figure}
\par      
In Fig.5. the 3 colored rectangular represents:
\begin{itemize}
\item Red rectangular zone: The heavy traffic load zone.
\item Green rectangular zone: The traffic bottleneck zone.
\item Blue rectangular zone: The main exit of the cloverleaf system.
\end{itemize}
\par
During running the scenario, controlled by different $m$ values, cars (faster vehicles), trucks (slower vehicles) and buses (mid-speed vehicles) are generated rapidly from each possible entry. Although, there are 7 exits, the one on the left edge is design to be the main exit and 60\% of the total vehicles try to leave the overpass through this exit.
\begin{figure}[bhtp]
      \centering
     
      \includegraphics[scale=0.25]{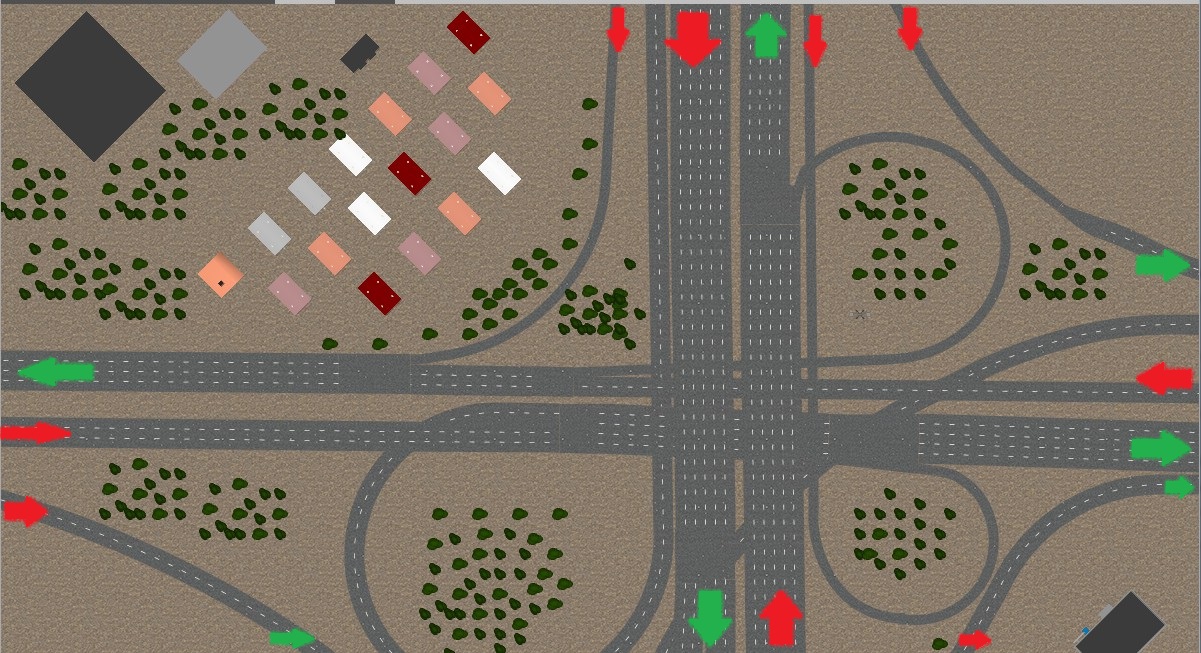}
      \caption{The entrance and exit of the cloverleaf interchange system}
      \label{figurelabel}
   \end{figure}
\par      
A density map is real-time generated to give us a perspective about the working status at the dynamic routing level. The color gradient from green to red reveals the light to heavy load transformation on that road section.
If we directly test the scenario with ordinary algorithm that each vehicle just tries to run as fast as tit can. Ignoring the exit owns 4 lanes, no one drives a long detour to use the 2 lanes on the outer edge until a quick congestion happens. 
\par 
 \begin{figure}[bpth]

       \centering\includegraphics[scale=0.25]{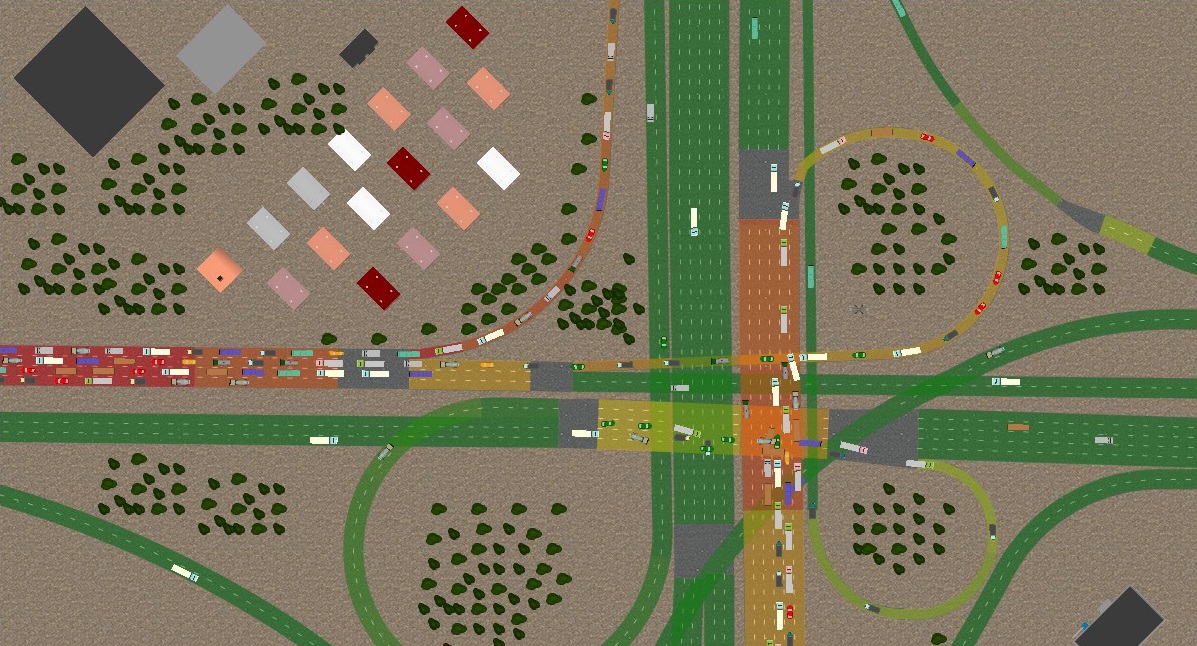}
       \caption{The cloverleaf interchange system get congestion after some time}
       \label{figurelabel}
   \end{figure}
Our proposed SPH based dynamic path planning strategy can lead the vehicles to pass through the flyover and leave through the outer side of the exit from initial. As far as we simulated, the congestion rarely happens under the guide of dynamic path planning.
In 10 consecutive runs, the system time of the left exit get stuck (red zone under density view) under 2 strategy is:
\begin{table}[h]
\caption{left exit congested time}
\label{table 1}
\begin{center}
\begin {tabular}{c c c} 
 Runs & Congested Time under Blind & Dynamic Routing\\
\hline
 1 & $829$ & $N/A$ \\
 2 & $783$ & $N/A$ \\
 3 & $779$ & $N/A$ \\
 4 & $802$ & $N/A$ \\
 5 & $774$ & $3374$ \\
 6 & $833$ & $N/A$ \\
 7 & $801$ & $N/A$ \\
 8 & $782$ & $N/A$ \\
 9 & $791$ & $N/A$ \\
 10 & $842$ & $N/A$ \\  
 11 & $766$ & $N/A$ \\
 12 & $802$ & $N/A$ \\
 13 & $864$ & $N/A$ \\
 14 & $848$ & $N/A$ \\
 15 & $761$ & $N/A$ \\
 16 & $742$ & $3452$ \\
 17 & $797$ & $N/A$ \\
 18 & $863$ & $N/A$ \\ 
 19 & $952$ & $N/A$ \\ 
 20 & $755$ & $N/A$ \\ 
\hline
\end{tabular}
\end{center}
\end{table}
\newline
\par 
If no congestion happens within 3600s, $N/A$ is placed. The above table can tell the huge difference of whether applying the dynamic routing in system level. Our dynamic routing method performs quite satisfying in a complex environment testing.
\section{Conclusion}
We established a dynamic routing method that can cater to drivers' preferences and optimize the traffic load of the entire traffic flow according to MAS and SPH algorithms. Moreover, as a discrete control method, the drivers' actions will not cause failure or roll-back of the routing system. We will process real experiments with our dynamic routing method for real traffic system in the future.





\begin{thebibliography}{99}

\bibitem{c1} Nagel, Kai, and Michael Schreckenberg. "A cellular automaton model for freeway traffic." Journal de physique I 2.12 (1992): 2221-2229.
\bibitem{c2} Stentz, Anthony. "Optimal and efficient path planning for partially known environments." Intelligent Unmanned Ground Vehicles. Springer, Boston, MA, 1997. 203-220.
\bibitem{c3} Cremer, M., and Markos Papageorgiou. "Parameter identification for a traffic flow model." Automatica 17.6 (1981): 837-843.
\bibitem{c4} May, Adolf D. Traffic flow fundamentals. 1990.
\bibitem{c5} Daganzo, Carlos F. "The cell transmission model: A dynamic representation of highway traffic consistent with the hydrodynamic theory." Transportation Research Part B: Methodological 28.4 (1994): 269-287.
\bibitem{c6} Pac, Muhammed R., Aydan M. Erkmen, and Ismet Erkmen. "Control of robotic swarm behaviors based on smoothed particle hydrodynamics." 2007 IEEE/RSJ International Conference on Intelligent Robots and Systems. IEEE, 2007.
\bibitem{c7} Pimenta, Luciano CA, et al. "Swarm coordination based on smoothed particle hydrodynamics technique." IEEE Transactions on Robotics 29.2 (2013): 383-399.
\bibitem{c8} Sabattini, Lorenzo, et al. "Implementation of coordinated complex dynamic behaviors in multirobot systems." IEEE Transactions on Robotics 31.4 (2015): 1018-1032.
\bibitem{c9} Müller, Matthias, David Charypar, and Markus Gross. "Particle-based fluid simulation for interactive applications." Proceedings of the 2003 ACM SIGGRAPH/Eurographics symposium on Computer animation. Eurographics Association, 2003.
\bibitem{c10} Monaghan, Joe J. "Smoothed particle hydrodynamics." Annual review of astronomy and astrophysics 30.1 (1992): 543-574.
\bibitem{c11} Savino, Heitor Judiss, F. O. Souza, and Luciano CA Pimenta. "Consensus with guaranteed convergence rate of high-order integrator agents in the presence of time-varying delays." International Journal of Systems Science 47.10 (2016): 2475-2486.
\bibitem{c12} Sabattini, Lorenzo, Cristian Secchi, and Cesare Fantuzzi. "Coordinated dynamic behaviors for multirobot systems with collision avoidance." IEEE transactions on cybernetics 47.12 (2017): 4062-4073.
\bibitem{c13} Pimenta, Luciano CA, et al. "Control of swarms based on hydrodynamic models." 2008 IEEE International Conference on Robotics and Automation. IEEE, 2008.
\bibitem{c14} M. A. Hsieh and V. Kumar, “Pattern generation with multiple robots,” in Proc. IEEE Int. Conf. Robot. Autom., Orlando, FL, May. 2006, pp. 2442–2447
\bibitem{c15} C. I. Connolly, J. B. Burns, and R. Weiss, “Path planning using Laplace’s
equation,” in Proc. IEEE Int. Conf. Robot. Autom., May 1990, pp. 2102–
2106
\bibitem{c16} Waydo, Stephen, and Richard M. Murray. "Vehicle motion planning using stream functions." 2003 IEEE International Conference on Robotics and Automation (Cat. No. 03CH37422). Vol. 2. IEEE, 2003.
\bibitem{c17} May, Adolf D. Traffic flow fundamentals. 1990.
\bibitem{c18} Lebacque, Jean-Patrick. "First-order macroscopic traffic flow models: Intersection modeling, network modeling." Transportation and Traffic Theory. Flow, Dynamics and Human Interaction. 16th International Symposium on Transportation and Traffic TheoryUniversity of Maryland, College Park. 2005.
\bibitem{c19} D. Lipinski and K. Mohseni, “A master-slave fluid cooperative control algorithm for optimal trajectory planning,” in Proc. IEEE Int. Conf. Robot. Autom., Shangai, China, May 2011, pp. 3347–3351.
\bibitem{c20} Warren, Charles W. "Global path planning using artificial potential fields." Proceedings, 1989 International Conference on Robotics and Automation. Ieee, 1989.
\bibitem{c21} Barraquand, Jerome, Bruno Langlois, and J-C. Latombe. "Numerical potential field techniques for robot path planning." IEEE transactions on systems, man, and cybernetics 22.2 (1992): 224-241.
\bibitem{22} Ehmke, J., \& Mattfeld, D. (2011). Integration of information and optimization models for vehicle routing in urban areas. In The state of the art in the European Quantitative Oriented Transportation and Logistics Research 14th Euro Working Group on Transportation, 26th Mini Euro Conference, 1st European Scientific Conference on Air Transport (Vol. 20, pp. 110–119).
\bibitem{23} COST. (2011). Towards autonomic road transport support systems. Retrieved from http://-www.cost.esf.org/-domains actions/-tud/-Actions/-TU1102?


\end{thebibliography}
\end{document}